\documentclass{article}

\usepackage{PRIMEarxiv}

\usepackage[utf8]{inputenc} 
\usepackage[T1]{fontenc}    
\usepackage{hyperref}       
\usepackage{url}            
\usepackage{booktabs}       
\usepackage{amsfonts}       
\usepackage{nicefrac}       
\usepackage{microtype}      
\usepackage{lipsum}
\usepackage{graphicx}

\usepackage{colortbl}
\usepackage{pgfplots}
\usepackage{pgfplotstable}

\usepackage{amssymb}
\usepackage{algorithm}
\usepackage{algorithmicx}
\usepackage{algpseudocode}

\usepackage{amsmath}
\DeclareMathOperator*{\argmax}{arg\,max}

\usepackage{xcolor}
\definecolor{orchid}{RGB}{218,112,214}
\definecolor{turquoise}{RGB}{64,224,208}
\definecolor{navy}{RGB}{0,0,128}
\definecolor{olive}{RGB}{128,128,0}
\definecolor{maroon}{RGB}{128,0,0}
\definecolor{gold}{RGB}{255,215,0}
\definecolor{silver}{RGB}{192,192,192}
\definecolor{turquoise}{RGB}{64,224,208}
\definecolor{lavender}{RGB}{230,230,250}
\definecolor{coral}{RGB}{255,127,80}
\definecolor{indigo}{RGB}{75,0,130}
\definecolor{salmon}{RGB}{250,128,114}
\definecolor{beige}{RGB}{245,245,220}
\definecolor{skyblue}{RGB}{135,206,235}
\definecolor{limegreen}{RGB}{50,205,50}
\definecolor{peach}{RGB}{255,218,185}
\definecolor{mintgreen}{RGB}{152,251,152}
\definecolor{slategray}{RGB}{112,128,144}
\definecolor{brickred}{RGB}{178,34,34}
\definecolor{darkpurple}{RGB}{148,0,211}
\definecolor{forestgreen}{RGB}{34,139,34}
\definecolor{midnightblue}{RGB}{25,25,112}
\definecolor{rubyred}{RGB}{224,17,95}
\definecolor{steelblue}{RGB}{70,130,180}
\definecolor{sandybrown}{RGB}{244,164,96}
\definecolor{hotpink}{RGB}{255,105,180}
\definecolor{lightgray}{RGB}{211,211,211}
\definecolor{babyblue}{RGB}{137,207,240}
\definecolor{ivory}{RGB}{255,255,240}
\definecolor{chartreuse}{RGB}{127,255,0}
\definecolor{orchid}{RGB}{218,112,214}
\definecolor{khaki}{RGB}{240,230,140}
\definecolor{periwinkle}{RGB}{204,204,255}
\definecolor{olivegreen}{RGB}{107,142,35}
\definecolor{chocolatebrown}{RGB}{210,105,30}
\definecolor{turquoiseblue}{RGB}{0,199,140}
\definecolor{plum}{RGB}{221,160,221}
\definecolor{slateblue}{RGB}{106,90,205}

\definecolor{area_0}{RGB}{221,190,169}
\definecolor{area_1}{RGB}{42,157,143}
\definecolor{area_2}{RGB}{233,196,106}
\definecolor{area_3}{RGB}{244,162,97}
\definecolor{area_4}{RGB}{169,222,249}
\definecolor{area_5}{RGB}{205,180,219}

\definecolor{deepgreen}{rgb}{0.0, 0.2, 0.0}

\usepackage{csvsimple}
\pgfplotsset{compat=1.17}

\usepackage{tikz}
\usepackage{collcell}
 
\newcommand*{\MinNumber}{0.1}%
\newcommand*{\MaxNumber}{0.9}%
 
\newcommand{\ApplyGradient}[1]{%
        \pgfmathsetmacro{\PercentColor}{100.0*(#1-\MinNumber)/(\MaxNumber-\MinNumber)}
        \hspace{-0.33em}\colorbox{deepgreen!\PercentColor!white}{}
}
 
\newcolumntype{R}{>{\collectcell\ApplyGradient}c<{\endcollectcell}}

\usetikzlibrary{shapes.geometric}
\usetikzlibrary{shapes.symbols}
\usetikzlibrary{calc}
\usetikzlibrary{arrows, positioning}
\usetikzlibrary{positioning, fit, arrows.meta, backgrounds, shadows}
\usetikzlibrary{decorations.pathreplacing}

\tikzset{
    module/.style={%
        draw, rounded corners,
        minimum width=#1,
        minimum height=7mm,
        font=\sffamily
        },
    module/.default=2cm,
    >=LaTeX
}

\title{Adaptive Learning Path Navigation Based on Knowledge Tracing and Reinforcement Learning
}

\usepackage{authblk}

\author{
  \begin{tabular}[t]{@{}c@{}}
    Jyun-Yi Chen \\
    National Taiwan Normal University \\
    \texttt{61175017h@ntnu.edu.tw}
  \end{tabular}
  \and
  \begin{tabular}[t]{@{}c@{}}
    Saeed Saeedvand \\
    National Taiwan Normal University \\
    \texttt{saeedvand@ntnu.edu.tw}
  \end{tabular}
  \and
  \begin{tabular}[t]{@{}c@{}}
    I-Wei Lai \\
    National Taiwan Normal University \\
    \texttt{iweilai@ntnu.edu.tw}
  \end{tabular}
}

\begin{document}
\maketitle

\begin{abstract}

This paper introduces the Adaptive Learning Path Navigation (ALPN) system, a novel approach for enhancing E-learning platforms by providing highly adaptive learning paths for students. The ALPN system integrates the Attentive Knowledge Tracing (AKT) model, which assesses students' knowledge states, with the proposed Entropy-enhanced Proximal Policy Optimization (EPPO) algorithm. This new algorithm optimizes the recommendation of learning materials. By harmonizing these models, the ALPN system tailors the learning path to students' needs, significantly increasing learning effectiveness. Experimental results demonstrate that the ALPN system outperforms previous research by 8.2\% in maximizing learning outcomes and provides a 10.5\% higher diversity in generating learning paths. The proposed system marks a significant advancement in adaptive E-learning, potentially transforming the educational landscape in the digital era.

\end{abstract}


\section{Introduction}
Integrating pedagogy and technology has brought significant transformations in education, with E-learning systems emerging as accessible, cost-effective, collaborative, and flexible alternatives to classroom-based education \cite{Accessibility, Flexibility, CostReduce, Collaboration}. E-learning systems enable students to access a wealth of learning materials for self-directed learning at their convenience. Considering these advantages, exploring strategies to enhance E-learning systems further and optimizing their effectiveness becomes crucial. One promising approach involves offering students well-designed learning paths on these systems. With the support of learning paths, students have a lower chance of encountering learning disorientation and cognitive overload, which improves their learning efficiency \cite{LearningDisorientation, CognitiveOverload1, CognitiveOverload2, CognitiveOverload3}.

A learning path constitutes an automatically curated sequence of learning materials to ensure the student has all the required knowledge to achieve their learning goal \cite{DefinitionLP}. Depending on the student's requirements, learning materials can be courses, topics, or learning objects \cite{LearningObject}. The methods of selecting and organizing learning materials to form a learning path have been studied for some time. Traditional E-learning systems commonly utilize standardized methods to provide fixed learning paths \cite{Hierarchical, Spiral, Ontology, LinearProgression, Thematic, Prerequisite}. However, these methods may make some students feel under-challenged or overwhelmed, as each student has a different learning goal and knowledge background \cite{ReduceEfficiency}. Hence, providing adaptive learning paths tailored to individuals has become a pressing concern.

In recent years, researchers have proposed different approaches for adaptive learning path planning using various personalization parameters. These parameters include learning goals \cite{LPbyCourses, Parameter1_Goal_Reward}, time limitations \cite{Parameter4_Time, Parameter5_Time}, knowledge backgrounds \cite{Parameter2_KB_Subjective, Parameter3_KB_Subjective}, etc. The learning goals can be deadline-driven or mastery-driven, determined by if the students have time limitations. The knowledge backgrounds refer to the knowledge level of students before they engage with the learning paths. When an adaptive learning path planning approach can accurately assess students' knowledge backgrounds in advance, it can recommend appropriate learning materials based on their deficiencies. However, as the number of students and learning materials on E-learning systems continues to grow, maintaining or improving the system's ability to diagnose students' learning progress becomes challenging \cite{LPsurvey}. To address this problem, developing scalable methods stands as a permanent solution.

This paper proposes an Adaptive Learning Path Navigation (ALPN) system that recommends learning materials to students according to their knowledge states, i.e., the mastery level of concepts in a subject. The system employs a Knowledge Tracing (KT) model to quantify students' current knowledge states \cite{KTsurvey}, followed by a decision-making model that recommends tailored learning materials. As students complete the learning materials, the KT model updates their knowledge states by assessing their responses. This iterative recommendation and knowledge state updating process continues until students achieve their learning goals. Throughout this process, the learning materials recommended by the decision-making model form the students' learning paths. By considering the dynamics of a student's knowledge state during the learning process, the proposed ALPN system generates highly adaptive learning paths. In addition, we implement two models using deep learning techniques to enhance scalability, enabling our system to maintain or improve its performance when facing more diverse students and learning materials \cite{DL, DRL}.

KT estimates students' knowledge states and predicts their future performance according to their learning records within an E-learning system \cite{BKT}. Utilizing deep learning techniques, KT has achieved significant advancements in learning diagnosis \cite{DKT}. Our system employs an attention-based model, Attentive Knowledge Tracing (AKT), to offer reliable assessments of students' learning progress \cite{Attention, AKT}. For another part, we propose an Entropy-enhanced version of Proximal Policy Optimization (PPO) called EPPO for learning material recommendations \cite{PPO}. EPPO demonstrates superior performance in optimizing the student's learning outcome compared to vanilla PPO in our task by incorporating enhanced exploratory capabilities. With the combination of  AKT and EPPO, our ALPN system can create effective learning paths.

In the experiments, we compared our proposed ALPN system with the Knowledge Tracing based Knowledge Demand Model (KT-KDM), currently the method most similar to ours in learning path recommendation research \cite{KTKDM}. Regarding maximizing students' learning outcomes, our ALPN system outperformed KT-KDM on average by 8.2\%. Moreover, the ALPN system exhibited a 10.5\% higher diversity in generating learning paths than KT-KDM. We also conducted additional analyses to demonstrate the performance of the ALPN system in various aspects.

In a nutshell, the proposed ALPN system integrates AKT and EPPO models, offering adaptive learning paths to students. The contributions of this work are listed below:

\begin{enumerate}{}
    \item Proposing the EPPO, an algorithm outperforming conventional PPO in learning path recommendation.

    \item Developing a system that can improve students' learning outcomes by 8.2\% more than the previous similar approach.

    \item Applying a KT model based on the attention mechanism to assess a student's knowledge state thus makes the generated adaptive learning paths more reliable.

    \item Creating a framework that can continuously update a student's knowledge state, enabling the system to consistently recommend the most suitable learning materials based on their needs.
\end{enumerate}

The paper presents an organization: Section 2 overviews the relevant existing works, Section 3 describes the mechanism of KT, and Section 4 elaborates on the learning path planning method. Next, Section 5 introduces the dataset and presents the experimental results. Finally, Section 6 concludes this paper and proposes directions for future research.

\section{Related Work}
This section reviews relevant prior work in knowledge tracing and reinforcement learning.

\subsection{Knowledge Tracing}

KT models have two categories: traditional knowledge tracing models and deep learning-based knowledge tracing models.

\begin{itemize}
  \item Item Response Theory: Item Response Theory (IRT) is a popular probabilistic framework in psychometrics that assesses latent traits based on individuals' responses to a collection of items \cite{IRT}. Although IRT demonstrates robustness across diverse domains, it falls short in capturing the dynamic nature of learning due to the assumption that the latent trait (e.g., a student's knowledge level) remains constant over time. Furthermore, IRT does not explicitly model the sequence of item interactions, a critical aspect for comprehending learning processes.

  \item Bayesian Knowledge Tracing: Bayesian Knowledge Tracing (BKT) addresses the limitations of IRT by representing a student's learning process as a hidden Markov model \cite{BKT}. This modeling technique monitors an individual's proficiency in knowledge components over time, considering the sequential order of item interactions. Nonetheless, BKT permits only a binary representation of knowledge mastery (i.e., learned or unlearned), indicating that its effectiveness warrants further enhancement.

  \item Deep Knowledge Tracing: Deep Knowledge Tracing (DKT) expands the application of deep learning in knowledge tracing, employing Recurrent Neural Networks (RNN) or Long Short-Term Memory (LSTM) to model individuals' dynamic knowledge acquisition \cite{DKT}. By leveraging neural network computations, DKT can represent a more extensive range of knowledge states. Consequently, it offers a nuanced and precise approximation of a student's learning progress, demonstrating the potential for adaptive learning path recommendations.
\end{itemize}

\subsection{Reinforcement Learning}

In recent years, Reinforcement Learning (RL) has emerged as a prominent field within artificial intelligence and machine learning. It aims to develop algorithms that allow agents to learn optimal policies through interactions with their environment, aiming to maximize cumulative rewards \cite{SuttonBarto2018}. Classic RL algorithms, such as Q-learning and SARSA, rely on tabular representations of state-action values, which may suffer from the curse of dimensionality in cases with large state spaces \cite{Q_learning, SARSA}. This limitation has led to exploring function approximation techniques to generalize across states and actions, such as linear function approximation and tile coding \cite{ResidualAlgorithms, tile_coding}. However, these methods often fail to scale well or learn efficiently in complex environments with high-dimensional states and action spaces.

Deep Reinforcement Learning (DRL) was introduced to overcome traditional RL methods' limitations, integrating deep learning techniques with reinforcement learning algorithms. DRL leverages the power of deep neural networks to represent the value functions or policies, enabling the learning of complex features and representations in high-dimensional spaces \cite{DQN}. One breakthrough example is the Deep Q-Network (DQN) algorithm, which successfully learned to play a wide range of Atari games directly from raw pixel inputs, outperforming many previous methods \cite{Atari}. DRL has also demonstrated remarkable success in domains such as robotic control \cite{EndToEndTraining}, natural language processing \cite{NMT_JointlyLearning}, and game-playing \cite{MasteringGo}. Unlike traditional RL methods, DRL offers significant improvements in scalability and generalization, enabling agents to learn more effectively in complex, high-dimensional environments.

\section{Knowledge Tracing Model}

To ensure that each learning material provided by the system is most effective for the students, we need to track their learning progress as they follow the learning path. In this regard, we employ the KT technique to assess the student's knowledge state continuously. We describe the central concepts of KT in this section and briefly introduce the KT approach we have adopted.

\subsection{Problem Definition}

KT quantifies the likelihood that a student will correctly answer specific exercises based on the student's learning history. Given a set of exercise indices $\mathcal E = \{e_{1}, e_{2}, \ldots, e_{J}\} \in \mathbb{N}$ and a student's historical interaction sequence $\mathcal I = (I_{1}, I_{2}, \ldots, I_{t})$ ordered by time. For each interaction log $I_{t}=(e_{t}, c_{t})$, $e_{t} \in \mathcal E$ denotes the exercise index that the student answered at time $t$, and $c_{t} \in \{0, 1\}$ represents the correctness of the answer. Using these interaction logs as input, the KT model outputs a vector $\mathbf s_{t}=[s_{1, t}, s_{2, t}, ..., s_{J, t}]^\top$ that represents the student's knowledge state at time $t$. Each element in the vector $\mathbf s_{t}$ represents the probability of the student correctly answering a specific exercise in $\mathcal E$:

\begin{equation}
s_{j,t} = \mathbb P[c_{j} = 1|I_{1}, I_{2}, ..., I_{t-1}, e_{j}].
\end{equation}

We consider the knowledge state to represent the student's knowledge background and utilize it as a personalization parameter to influence the recommendation of the subsequent learning path.

\subsection{Applied Scheme}

In this study, we are motivated by the AKT scheme \cite{AKT} to employ the KT model for assessing students' knowledge state. The AKT employs an exercise self-attentive encoder to convert each input exercise index into a contextualized representation. Then, it evaluates the student's acquired knowledge from past interactions by using a knowledge acquisition self-attentive encoder. Following the knowledge acquisition step, the AKT filters the previously acquired knowledge to extract the relevant knowledge associated with the current exercise utilizing a single attention-based knowledge retriever. Finally, it evaluates the probability of the student correctly answering a specific exercise using the extracted knowledge.

\section{Learning Path Navigation System}

The proposed ALPN system employs a pre-trained AKT model as the environment, allowing the decision-making model, i.e., the agent, to explore diverse learning paths. When the student follows a learning path, multiple state-action-reward tuples will be generated over time, collectively forming a trajectory:

\begin{equation}
\tau = \{(\mathbf s_{1}, a_{1}, r_{1}), (\mathbf s_{2}, a_{2}, r_{2}), \ldots, (\mathbf s_{T}, a_{T}, r_{T})\},
\end{equation}

\noindent where $\mathbf s_{t}$ denotes the student's knowledge state, $a_{t}$ represents the exercise index recommended by the agent, and $r_{t}$ signifies the immediate reward that the agent received after the student completes the exercise. We assign the maximum length of a single learning path to $\textit{T}_\text{max}$, such that $\textit{T} \leq \textit{T}_\text{max}$. Notably, $\mathbf s_{1}$ indicates the knowledge background the AKT evaluates before the student engages with the learning path.

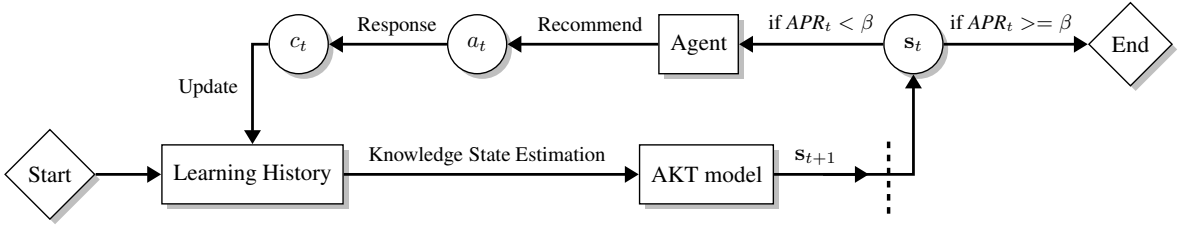
\begin{figure*}[t!]
    \centering
    \resizebox{0.95\textwidth}{!}{
    \begin{tikzpicture}[
    node distance = 12mm and 18mm,
        arr/.style = {-Triangle, very thick},
        box/.style = {rectangle, draw, semithick,
                     minimum height=9mm, minimum width=9mm,
                     fill=white, drop shadow},
                            ]
    \node (n1) [box, inner sep=5pt] {Learning History};
    \node (n2) [box, inner sep=5pt, right=of n1, xshift=27mm] {AKT model};
    \node (n3) [circle, above right=of n2, draw, semithick, minimum size=9mm, fill=white, drop shadow] {$\mathbf s_{t}$};
    \node (n4) [diamond, right=of n3, xshift=4mm, draw, semithick, minimum size=9mm, fill=white, drop shadow] {End};
    \node (n5) [box, inner sep=5pt, xshift=-4mm, left=of n3] {Agent};
    \node (n6) [circle, left=of n5, xshift=-5mm, draw, semithick, minimum size=9mm, fill=white, drop shadow] {$a_{t}$};
    \node (n7) [circle, left=of n6, draw, semithick, minimum size=9mm, fill=white, drop shadow] {$c_{t}$};
    \node (n8) [diamond, left=of n1, xshift=8mm, draw, semithick, minimum size=9mm, fill=white, drop shadow] {Start};

    \draw[arr] (n1) -- node[pos=0.485, above] {\small Knowledge State Estimation} (n2);
    \draw[arr] (n2) -| node[pos=0.15, above] {$\mathbf s_{t+1}$} (n3);
    \draw[arr] (n3) -- node[pos=0.45, above, xshift=0.25ex, yshift=0.2ex] {\small if $\textit{APR}_t >= \beta$} (n4);
    \draw[arr] (n3) -- node[pos=0.43, above, yshift=0.2ex] {\small if $\textit{APR}_t < \beta$} (n5);
    \draw[arr] (n5) -- node[pos=0.45, above, yshift=0.4ex] {\small Recommend} (n6);
    \draw[arr] (n6) -- node[pos=0.42, above] {\small Response} (n7);
    \draw[arr] (n7) -| node[pos=0.72, left, xshift=-0.5ex] {\small Update} (n1);
    \draw[arr] (n8) -- (n1);

    \draw[arr] (8.7, 0) -- (9.4, 0);
    \draw[dash pattern=on 1mm off 1mm, line width=1.5pt] (9.7, -0.6) -- (9.7, 0.58);

    \end{tikzpicture}
    }
    \vspace{7pt} 
    \caption{The workflow of the proposed ALPN system.}
    \label{fig:framework}
\end{figure*}

In Figure \ref{fig:framework}, we present the conceptual view of our ALPN system framework. First, AKT analyzes the student's learning history and determines their initial knowledge state. Subsequently, the agent takes this knowledge state as input and recommends appropriate learning material (exercise) to the student. Upon completion of the exercise by the student, the system incorporates the interaction record into their learning history and reevaluates the knowledge state. This process continues until the student's knowledge level surpasses the predetermined learning goal. The following subsections will delve into the various components employed in the system's implementation.

\subsection{Probabilistic Formulation}

Our system's primary objective is to provide learning paths to students with various knowledge backgrounds, ensuring they achieve their learning goals. The optimal approach involves augmenting the probability of generating trajectories with the highest cumulative reward across heterogeneous initial knowledge states. Hence, we commence by defining the probability formulation:

\begin{equation}
\label{eq:4}
\mathbb P_{\theta}[\tau] = \mathbb P[\mathbf s_{1}] \prod_{t=1}^{T} \pi_{\theta}(a_{t}|\mathbf s_{t}) \mathbb P[c_{t}|\mathbf s_{t},a_{t}] \mathbb P[\mathbf s_{t+1}|\mathbf s_{t},a_{t},c_{t}],
\end{equation}

\noindent where $\mathbb P[\mathbf s_{1}]$ indicates the initial knowledge state distribution, and $\mathbb P[\mathbf s_{t+1}|\mathbf s_{t},a_{t},c_{t}]$ stands for the transition probability of knowledge state. The term $\mathbb P[c_{t}|\mathbf s_{t}, a_{t}]$ represents the Bernoulli distribution of the probability that the student correctly answers the exercise $a_{t}$, as determined by the AKT model.

\subsection{Learning Goal}

In our ALPN system, we set the primary learning goal for the student as maximizing their learning level. We define the student's learning level as the average pass rate ($\textit{APR}$) for all exercises. We can express the student's $\textit{APR}$ at time $t$ as:

\begin{equation}
\textit{APR}_t = \frac{1}{\left| \mathcal A \right|} \sum_{j=1}^{\left| \mathcal A \right|}s_{j,t}\ , \quad \forall j: s_{j,t} \in (0, 1),
\end{equation}

\noindent where $\left| \cdot \right|$ denotes the cardinality of a set, which represents the number of elements in the set. Specifically, $\left| \mathcal A \right|$ represents the number of available exercises in the E-learning system and denotes the action space size for the agent. $s_{j,t}$ indicates the probability of a student answering the $j$-th exercise in the exercise set correctly at time $t$.

Upon completion of an exercise that the agent recommends, the student's $\textit{APR}_t$ undergoes comparison with a predetermined threshold $\beta$. When the $\textit{APR}_t$ exceeds or equals $\beta$, the system deems the student to have accomplished the learning goal. The $\beta$ does as a hyperparameter, allowing for adjustments as required.

\subsection{Reward Function}

We develop the reward function to align with the student's learning goal. It can evaluate the quality of the ALPN agent's actions and guides its decision-making by reinforcing actions that yield desirable results. We emphasize the learning gain as a critical parameter to achieve this. Learning gain refers to the distinction between students' skills, competencies, content knowledge, and personal development at two different points in time \cite{LG}. It varies from learning outcomes, as learning gain compares performance at two-time points, whereas learning outcomes focus on knowledge level after finishing a learning session. Precisely, we calculate learning gain as $\textit{LG}_t = \textit{APR}_t - \textit{APR}_{t-1}$, where $\textit{APR}_t$ and $\textit{APR}_{t-1}$ indicate a student's current knowledge level and previous knowledge level.

This study also considers the distance between a student's current knowledge level and the learning goal, as an essential parameter. We define the distance parameter as $\textit{d}_t = \beta - \textit{APR}_t$, which acts as a factor influencing the reward signal's magnitude. As the distance parameter gets incorporated, dividing the learning gain by distance returns the central part of the reward signal, reflecting the challenge associated with further enhancing the student's knowledge level when approaching their learning goal.

Furthermore, we regard the diversity in learning materials (i.e., exercises) as essential within a single learning path. We apply a penalty parameter $\lambda$ into the reward function to ensure diversity, with the definition as:

\begin{equation}
\lambda = \dfrac{\textit{d}_1 \left| \mathcal A \right|}{\textit{T}_\text{max}},\quad \textit{d}_1 = \beta - \textit{APR}_1,
\end{equation}

\noindent where $\textit{d}_1$ denotes the initial distance between the student's knowledge level and the learning goal. $n_{j,t}$ indicates how many times the agent has recommended the $j$-th exercise in the current learning path at time $t$.

With these parameters, we formulate the reward function as follows:

\begin{equation}
r_{t} = \begin{cases}
\vspace{2mm}
\dfrac{\displaystyle \textit{LG}_t \left| \mathcal A \right|}{\displaystyle \textit{d}_t} - \lambda^{n_{j,t}} & \text{if } \textit{LG}_t \ge 0 \\
\dfrac{\displaystyle \textit{LG}_t \left| \mathcal A \right|}{\displaystyle \textit{d}_t} & \text{if } \textit{LG}_t < 0.
\end{cases}
\end{equation}

Deserving of mentioning exists that a more elevated initial knowledge level of the student (i.e., a smaller $\textit{d}_1$) corresponds to a less penalty parameter value. Reducing constraints on the agent allows for increased exploration opportunities, assessing the potential benefits of student review for specific exercises.

\subsection{Maximize Expected Reward}

The ALPN agent learns a target policy $\pi_{\boldsymbol{\theta}}(\cdot|\mathbf s_{t})$ that correlates the student's knowledge state $\mathbf s_{t}\in\mathcal S$ to a distribution over all actions $a\in\mathcal A$, with the objective of maximizing the expected cumulative rewards:

\begin{equation}
\bar{R}_{\theta} = \mathbb{E}_{\tau\sim\pi_{\mathbf \theta}}[R(\tau)] = \sum_{\tau} R(\tau) \mathbb P_{\theta}[\tau].
\end{equation}

In order to achieve maximization, we require the computation of its gradient:

\begin{equation}
\begin{aligned}
\nabla \bar{R}_{\theta} &= \sum_{\tau} R(\tau) \nabla \mathbb P_{\theta}[\tau] = \sum_{\tau} R(\tau) \mathbb P_{\theta}[\tau] \nabla \log \mathbb P_{\theta}[\tau] \\
                        &= \mathbb{E}_{\tau\sim\pi_{\mathbf \theta}} [R_{\tau} \nabla \log \mathbb P_{\theta}[\tau]].
\end{aligned}
\end{equation}

However, computing the specific expected value proves infeasible. Thus, we approximate it by sampling a sufficient number $N$ of trajectories:

\begin{equation}
\begin{aligned}
\nabla \bar{R}_{\theta} &\approx \frac{1}{N} \sum_{n=1}^{N} R(\tau^{n}) \nabla \log \mathbb P_{\theta}[\tau^{n}] \\
                        \label{eq:14}
                        &= \frac{1}{N} \sum_{n=1}^{N} \sum_{t=1}^{T} R(\tau^{n}) \nabla \log(\mathbb \pi_{\theta}(a_{t}^{n}|\mathbf s_{t}^{n}) \mathbb P[c_{t}^{n}|\mathbf s_{t}^{n}, a_{t}^{n}]),
\end{aligned}
\end{equation}

\noindent where the probability $\mathbb P[\mathbf s_{t+1}|\mathbf s_{t}, a_{t},c_{t}]$ in \eqref{eq:4} does not feature, as it stems from the trained AKT model in the environment. The knowledge state transition outcome exhibits determinism for a given pair $(\mathbf s_t, a_t, c_{t})$. Namely, $\log \mathbb P[\mathbf s_{t+1}|\mathbf s_{t},a_{t},c_{t}]$ reduces to $0$, rendering its inclusion in \eqref{eq:14} redundant.

With the gradient, we employ the gradient ascent algorithm to optimize the parameters $\theta$ of the policy network, resulting in the identification of the optimal policy $\pi_{\theta}^{*}(\cdot|\mathbf s_{t})$, which can maximize the students' long-term learning gains.

\subsection{Advantage Function}

From a practical implementation standpoint, when a student answers a recommended exercise correctly, the obtained learning gain is predominantly positive. This circumstance leads the agent to need help identifying the most beneficial exercises for the student in the long run, as it frequently receives reward signals greater than $0$. Consequently, we utilize the following advantage function for replacing the original $R(\tau)$:

\begin{equation}
\begin{aligned}
A_{\theta}(\mathbf s_{t}, a_{t}) &= \Sigma_{t'=t}^{T}(\gamma^{t'-t} r_{t'}) - V_{\theta}(\mathbf s_{t}) \\
                                 &= Q_{\theta}(\mathbf s_{t}, a_{t}) - V_{\theta}(\mathbf s_{t}),
\end{aligned}
\end{equation}

\noindent where $\gamma \in [0, 1]$ denotes the discount factor, which balances the immediate rewards with the potential long-term rewards resulting from a particular action.

Operating this advantage function allows the agent to ascertain how recommending a specific exercise proves more effective than suggesting other exercises in a given knowledge state.

\begin{figure*}[t!]
\centering
\scalebox{0.9}{
\begin{tikzpicture}[
    every node/.style={draw, circle, minimum size=0.5cm},
    arrow/.style={},
    every label/.style={draw=none, align=center},
    rectangleNode/.style={draw, rectangle, minimum width=4.3cm, minimum height=0.75cm, fill=blue!10},
    rectangleS/.style={draw, rectangle, fill=green!10},
    rectangleH/.style={draw, rectangle, fill=gray!10},
    rectangleA/.style={draw, rectangle, fill=purple!10},
    rectangleC/.style={draw, rectangle, fill=orange!10}
]

\draw[decorate,decoration={brace,amplitude=10pt}] (-1,-4.75) -- (-1,0);
\draw[decorate,decoration={brace,amplitude=10pt}] (13.4,0) -- (13.4,-3.6);

\node[rectangleS, minimum width=1.25cm, minimum height=6cm] (ss) at (0,-2.35) {};
\node[draw=none] (Stext) at (-1.65,-2.375) {$\mathbf s_{t}$};
\node[fill=white](s1) at (0,0) {$s_{1, t}$};
\node[fill=white, below=0.5cm of s1] (s2) {$s_{2, t}$};
\node[fill=white, below=0.5cm of s2] (s3) {$s_{3, t}$};
\node[draw=none, below=0.5cm of s3, rotate=90, yshift=0.35cm] (dots1) {\ldots};
\node[fill=white, below=1cm of s3] (s718) {$s_{J, t}$};

\node[rectangleH, minimum width=4cm, minimum height=4cm, xshift=-0.015cm] (hh) at (4,-2.25) {};
\node[draw=none] (Htext) at (4,-4.65) {\textit{Hidden layers}};
\node[fill=white, right=2cm of s1, yshift=-1cm] (h1) {};
\node[fill=white, below=0.5cm of h1] (h2) {};
\node[draw=none, below=0.5cm of h2, rotate=90, yshift=0.35cm] (dots2) {\ldots};
\node[fill=white, below=1cm of h2] (h256) {};
\node[fill=white, fill=white, right=2cm of h1] (hh1) {};
\node[fill=white, below=0.5cm of hh1] (hh2) {};
\node[draw=none, below=0.5cm of hh2, rotate=90, yshift=0.35cm] (dots22) {\ldots};
\node[fill=white, below=1cm of hh2] (hh256) {};

\node[right=8cm of s1] (p1) {};
\node[below=0.5cm of p1] (p2) {};
\node[below=0.5cm of p2] (p3) {};
\node[draw=none, below=0.5cm of p3, rotate=90, yshift=0.35cm] (dots3) {\ldots};
\node[below=1cm of p3] (p718) {};
\node[right=8cm of s718] (v1) {};

\node[rectangleA, minimum width=1.68cm, minimum height=4.3cm, xshift=-0.015cm] (actor) at (12.18,-1.77) {};
\node[draw=none] (Atext) at (12.18,0.75) {\textit{Actor}};
\node[draw=none] (pitext) at (14.5,-1.77) {$\pi_{\theta}(\cdot|\mathbf s_{t})$};
\node[draw=none, right=2.35cm of p1] (pp1) {$\pi_{\theta}(e_1|\mathbf s_{t})$};
\node[draw=none, right=2.35cm of p2] (pp2) {$\pi_{\theta}(e_2|\mathbf s_{t})$};
\node[draw=none, right=2.35cm of p3] (pp3) {$\pi_{\theta}(e_3|\mathbf s_{t})$};
\node[draw=none, right=3.5cm of dots3, rotate=90, xshift=-0.7cm] (dots33) {\ldots};
\node[draw=none, right=2.35cm of p718] (pp718) {$\pi_{\theta}(e_J|\mathbf s_{t})$};

\node[rectangleC, minimum width=1.37cm, minimum height=0.75cm, xshift=-0.015cm] (hh) at (12.02,-4.71) {};
\node[draw=none] (Ctext) at (13.35,-4.71) {\textit{Critic}};
\node[draw=none, right=2.35cm of v1] (v11) {$V_{\theta}(\mathbf s_{t})$};

\node[right=1.5cm of p3, yshift=-1.9cm, xshift=-0.37cm, rectangleNode, rotate=90] (softmax) {\textit{Softmax function}};
\node[draw=none, right=0.75cm of p1, minimum size=0.75cm] (a1) {};
\node[draw=none, right=0.75cm of p2, minimum size=0.75cm] (a2) {};
\node[draw=none, right=0.75cm of p3, minimum size=0.75cm] (a3) {};
\node[draw=none, right=0.75cm of p718, minimum size=0.75cm] (a718) {};

\foreach \from/\to in {s1/h1, s1/h2, s1/h256, s2/h1, s2/h2, s2/h256, s3/h1, s3/h2, s3/h256, s718/h1, s718/h2, s718/h256,
                       h1/hh1, h1/hh2, h1/hh256, h2/hh1, h2/hh2, h2/hh256, h256/hh1, h256/hh2, h256/hh256,
                       hh1/p1, hh1/p2, hh1/p3, hh1/p718, hh1/v1, hh2/p1, hh2/p2, hh2/p3, hh2/p718, hh2/v1, hh256/p1, hh256/p2, hh256/p3, hh256/p718, hh256/v1,
                       p1/a1, p2/a2, p3/a3, p718/a718,
                       a1/pp1, a2/pp2, a3/pp3, a718/pp718, v1/v11}
\draw[arrow] (\from) -- (\to);

\end{tikzpicture}}
\caption{The network architecture of the ALPN agent.}
\label{fig:agent}
\end{figure*}
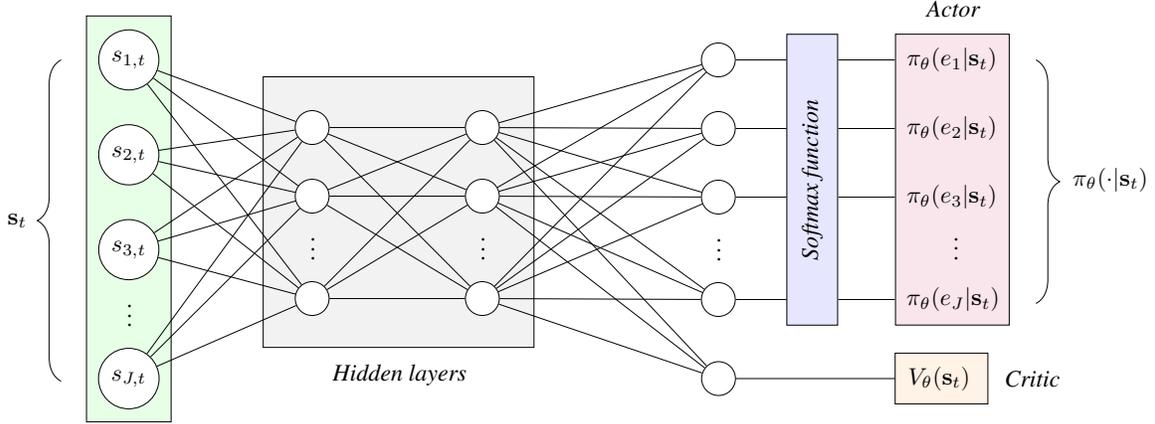

\subsection{Policy Gradient Method}

We adopt the Proximal Policy Optimization (PPO) algorithm \cite{PPO} as the foundation to construct our ALPN agent. PPO has been widely recognized for its exceptional reinforcement learning (RL) task performance, offering enhanced sample efficiency and training stability advantages. In this section, we present a comprehensive explanation of the PPO algorithm and describe its implementation within the framework of our proposed ALPN agent.

Figure \ref{fig:agent} illustrates the network architecture of the ALPN agent. The architecture comprises two parts: the actor and critic networks. The actor network determines the optimal policy that dictates the agent's actions in the AKT environment. It receives the input $\mathbf s_{t}$ and generates a probability distribution $\pi_{\theta}(\cdot|\mathbf s_{t})$ over the action space $\mathcal A$, guiding the agent's decisions. The primary objective of the actor entails maximizing the expected cumulative reward $\bar{R}_{\theta}$ when students follow their learning paths. For another part, the critic network serves as a value function estimator that calculates the value of the student's current state. Given the state $\mathbf s_{t}$ and policy $\pi_{\theta}$, this network takes the same input as the actor but outputs a scalar value $V_{\theta}(\mathbf s_{t})$ indicating the expected return. The critic aids in reducing the variance in the policy gradient estimate, thus enhancing the stability and convergence of the agent's learning process.

We utilize a replay buffer and the PPO's Clipped surrogate objective function to enhance the agent's sample efficiency. By incorporating a replay buffer, we can store past experiences, allowing the agent to learn from these experiences more data-efficiently. The Clipped surrogate objective function performs as a critical component in the PPO, addressing the central challenge of policy optimization \cite{DQN, DDPG, SuttonBarto2018}: balancing stable updates with sample efficiency. This function strives to prevent the policy from undergoing substantial updates, which could result in instability and suboptimal performance. We formulate the Clipped surrogate objective function as follows:

\begin{equation}
\begin{aligned}
L_{\theta^{k}}^{\textit{CLIP}}(\theta) &= \mathbb E_{(\mathbf s_{t}, a_{t}) \sim \pi_{\theta^{k}}}[ {\rm min}(\rho_{t}(\theta) A_{\theta^{k}}(\mathbf s_{t}, a_{t}),\ {\rm clip}(\rho_{t}(\theta),1-\epsilon,1+\epsilon)A_{\theta^{k}}(\mathbf s_{t}, a_{t}))],
\end{aligned}
\end{equation}

\noindent where $\theta^{k}$ means that we use the old policy $\pi_{\theta^{k}}$ to interact with the environment to collect transitions and compute advantage $A_{\theta^{k}}(\mathbf s_{t}, a_{t})$. Then, $\rho_{t}(\theta)$ indicates the probability ratio, which measures the relative likelihood of selecting a particular action under the new policy compared to the old policy:

\begin{equation}
\rho_{t}(\theta) = \frac{\pi_{\theta}(a_{t}|\mathbf s_{t})}{\pi_{\theta^{k}}(a_{t}|\mathbf s_{t})}.
\end{equation}

The probability ratio updates the policy, maintaining stability and ensuring consistency between the new and old policies. The PPO algorithm employs the Clipped surrogate objective function, which utilizes the probability ratio to balance stable policy updates and sample efficiency. Clipping the probability ratio within a specific range $[1-\epsilon,1+\epsilon]$ enables to avoid substantial policy updates that could cause instability and poor performance, where $\epsilon$ denotes a hyperparameter typically set to $0.2$.

We can further discuss the incorporation of two additional components in the objective function: the value function error term and the entropy bonus. These elements desire to address specific aspects of the learning process and enhance the overall performance of the ALPN system.

Incorporating the value function error decreases the mistake in value estimation from the critic, reducing the variability in the advantage function estimation process \cite{VF}. This enhanced precision, in turn, helps the agent make better decisions while navigating the learning path. Besides, when utilizing a neural network architecture that shares parameters between the actor and critic, operating an objective function that amalgamates the policy surrogate and a value function error term becomes essential.

The entropy bonus, counted to foster exploration and avert premature convergence to a suboptimal policy, encourages the agent to balance exploration and exploitation \cite{entropy1, entropy2}. This balance facilitates the discovery of novel strategies and elevates the agent's overall performance. Moreover, the entropy term helps the agent evade local optima and maintain learning progress throughout training.

Upon integrating these terms, the resulting objective function emerges, which the ALPN agent maximizes:

\begin{equation}
\begin{aligned}
L_{\theta^{k}}(\theta) &= \mathbb E_{(\mathbf s_{t}, a_{t}) \sim \pi_{\theta^{k}}}[L_{\theta^{k}}^{\textit{CLIP}}(\theta) - \frac{1}{2}L_{\theta^{k}}^{\textit{VF}}(\theta) + \alpha H[\pi_{\theta^{k}}](\mathbf s_{t})],
\end{aligned}
\end{equation}

\noindent where $L_{\theta^{k}}^{\textit{VF}}(\theta) = (V_{\theta^{k}}(\mathbf s_{t}) - V_{\theta}(\mathbf s_{t}))^2$ and $H[\pi_{\theta^{k}}](\mathbf s_{t})$ represents the entropy bonus. $\alpha$ denotes the temperature, which controls the trade-off between exploration and exploitation. A higher temperature value promotes exploration by increasing the randomness in the agent's recommendations. In contrast, a lower temperature value encourages exploitation by focusing more on the agent's current understanding.

\subsection{Entropy-Enhanced Proximal Policy Optimization}

Typically, when we incorporate the entropy bonus as part of the objective function, it is calculated based on the current policy. However, in this study, we explored an alternative approach. We stored the entropy computed by the policy at each timestep and the transitions in the replay buffer. In the objective function, we included the entropy from the buffer during policy updates. This approach allowed the agent to exhibit higher exploratory behavior in the early stages of training, facilitating the discovery of more suitable policies. We referred to the modified PPO with this adjustment as Entropy-enhanced Proximal Policy Optimization (EPPO). Through our experiments, we demonstrated the effectiveness of EPPO in providing optimal learning paths and improving the diversity of learning paths.

Finally, we provide a formal algorithmic description of the entire ALPN system comprising EPPO:

\begin{algorithm}
\caption{Adaptive Learning Path Navigation (ALPN) with EPPO}
\begin{algorithmic}[1]
\State Initialize agent parameters $\theta^{1}$
\State Initialize replay buffer $\mathcal{D}$
\For{$k=1,2,\ldots$}
    \State Input a student's learning history $\mathcal I$
    \State Input a student's initial knowledge state $\mathbf s_{1}$
    \For{$t=1,2,\ldots,\textit{T}_\text{max}$}
        \State Recommend exercise $a_{t}$ on policy $\pi_{\theta^{k}}(\cdot|\mathbf s_{t})$
        \State Set $c_{t}$ based on knowledge state $\mathbf s_{t}$
        \State Update learning history $\mathcal I$ with $a_{t}$, $c_{t}$
        \State Get $\mathbf s_{t+1}$ by $\mathcal I$ and AKT model
        \State Compute reward signal $r_{t}$
        \State Store transition $(\mathbf s_{t},a_{t},r_{t},\mathbf s_{t+1})$ and entropy bonus $H[\pi_{\theta^{k}}](\mathbf s_{t})$ in $\mathcal{D}$
        \If{$\textit{APR}_t \ge \beta$}
            \State \textbf{break}
        \EndIf
    \EndFor
    \State Sample transitions and entropy bonuses from $\mathcal{D}$
    \State Estimate advantages $A_{\theta^{k}}(\mathbf s_{1}, a_{1}),A_{\theta^{k}}(\mathbf s_{2}, a_{2}),\ldots,A_{\theta^{k}}(\mathbf s_{T}, a_{T})$
    \State Compute objective function $L_{\theta^{k}}(\theta)$
    \State Update policy parameters $\theta^{k+1} = \argmax_{\theta} L_{\theta^{k}}(\theta)$
\EndFor
\end{algorithmic}
\end{algorithm}

\section{Experiments}

In this section, we will examine the empirical evaluation of our proposed ALPN system. The section begins with introducing the dataset employed to train the AKT model within our system. Next, we will elucidate the evaluation methodology employed to assess the system's performance and present the agent's training process results.

Subsequently, we will provide additional analyses, including insights into the changes in the learning path length throughout the training process and an exploration of the diversity within the generated learning paths. These analyses will contribute to a more profound understanding of the ALPN system's efficacy and potential implications for adaptive learning.

\subsection{Dataset}

In our research, we train the AKT model within the ALPN system utilizing the Junyi Academy Math Practicing Log (Junyi) dataset \cite{junyi}. Comprising 25,925,922 interaction logs from 247,606 real-world students across 722 different math exercises, this dataset presents an exhaustive set of learning interactions. Each dataset exercise is designated with topic and area information, with 40 varied topics per exercise, such as absolute value, circle properties, and fractions. Areas represent broader categories, each encompassing several topics, and seven distinct areas are included, e.g., arithmetic, logic, and algebra. This dataset is open-sourced by Junyi Academy, one of Taiwan's leading online education platforms.

\subsection{Learning Outcome}

We evaluate the performance of our system by concentrating on the student's learning outcomes $\textit{APR}_T$ after completing the learning paths provided by the agent. This performance metric offers a practical insight into the agent's capability to generate adaptive learning paths that effectively improve students' knowledge states.

In the following experiments, we refer to the ALPN system that utilizes EPPO as ALPN-EPPO, and compare it with various variants of ALPN and a baseline:

\begin{itemize}
  \item ALPN-PPO: ALPN-PPO is the ALPN system that uses vanilla PPO for learning material recommendations.

  \item ALPN-A2C: ALPN-A2C represents the ALPN system that applies the Advantage Actor-Critic (A2C) algorithm.

  \item KT-KDM: KT-KDM, similar to our proposed ALPN system, employs the Deep Knowledge Tracing (DKT) method to track students' knowledge and utilizes A2C to provide learning paths \cite{KTKDM}.
\end{itemize}

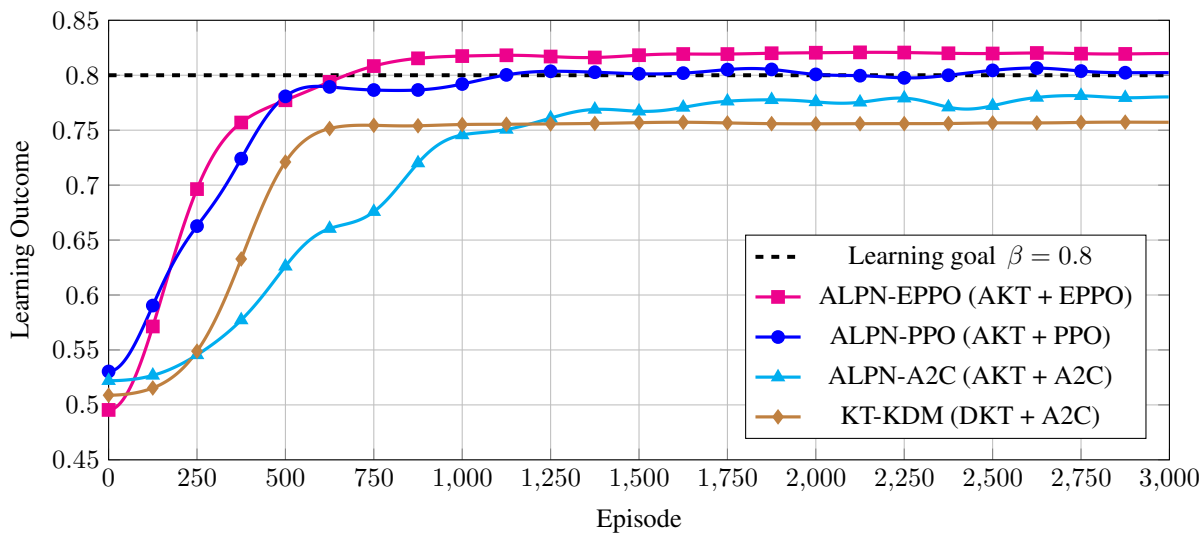
\begin{figure}[b]
\centering
\begin{tikzpicture}
\begin{axis}[
  xlabel={Episode},
  ylabel={Learning Outcome},
  xmin=0, xmax=3000,
  ymin=0.45, ymax=0.85,
  grid=major,
  legend entries={Learning goal\, $\beta=0.8$,\, ALPN-EPPO (AKT + EPPO), ALPN-PPO (AKT + PPO), ALPN-A2C (AKT + A2C), KT-KDM (DKT + A2C)},
  legend style={at={(0.6,0.28)}, anchor=west, row sep=0.1cm},
  width=0.95\textwidth,
  height=0.45\textwidth,
  xtick={0,250,...,3000},
  ytick={0.45,0.5,...,0.85}
]
\addplot[black, dashed, line width=1.5pt] coordinates {(0,0.8) (3000,0.8)};
\addplot[color=magenta, mark=square*, mark repeat=125, line width=1.2pt] table [x expr=\coordindex, y=ALPN-EPPO, col sep=comma] {score.csv};
\addplot[color=blue, mark=*, mark repeat=125, line width=1.2pt] table [x expr=\coordindex, y=ALPN-PPO, col sep=comma] {score.csv};
\addplot[color=cyan, mark=triangle*, mark repeat=125, line width=1.2pt] table [x expr=\coordindex, y=ALPN-A2C, col sep=comma] {score.csv};
\addplot[color=brown, mark=diamond*, mark repeat=125, line width=1.2pt] table [x expr=\coordindex, y=KT-KDM, col sep=comma] {score.csv};
\end{axis}
\end{tikzpicture}
\caption{The students' learning outcomes during the agents' training process.}
\label{figure:score}
\end{figure}

We compared ALPN-EPPO with other systems in terms of their performance during the training process over 3000 episodes. In each episode, a student was randomly sampled as the target for learning path recommendation, and each student had different learning records. With all students having a learning goal set to $0.8$, the result is shown in Figure~\ref{figure:score}. The figure shows that the ALPN-EPPO system is the most effective and robust in helping students achieve their learning goals. In comparison, the ALPN systems using general PPO or A2C and KT-KDM exhibit weaker performance. Additionally, it is worth noting that despite both KT-KDM and ALPN-A2C using the A2C algorithm for their decision-making models to recommend learning materials, there is a significant difference in convergence speed and stability between them. This discrepancy is attributed to the differences in the learning diagnosis models employed by the two systems, i.e., the differences between DKT and AKT.

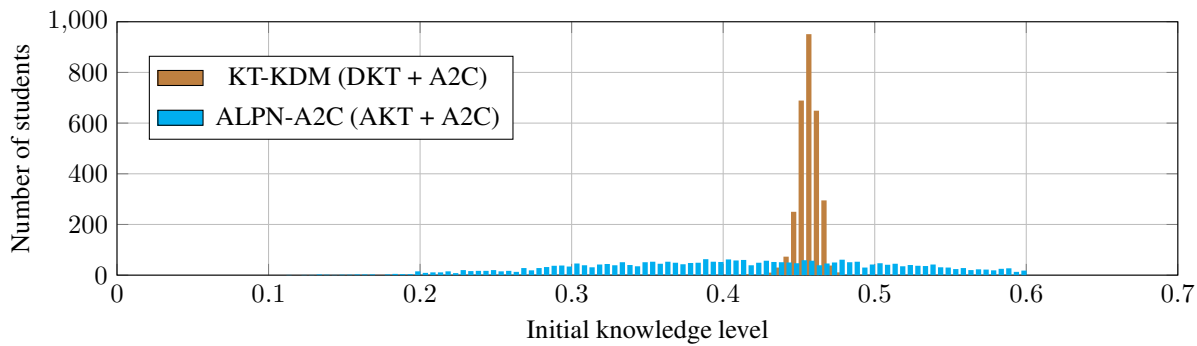
\begin{figure}[t]
    \centering
    \begin{tikzpicture}
    \begin{axis}[
        ybar,
        bar width=2pt,
        xlabel={Initial knowledge level},
        ylabel={Number of students},
        xmin=0, xmax=0.7,
        ymin=0, ymax=1000,
        grid=major,
        legend entries={\ KT-KDM (DKT + A2C),\ ALPN-A2C (AKT + A2C)},
        legend style={at={(0.03,0.7)}, anchor=west, row sep=0.1cm},
        area legend,
        width=0.95\textwidth,
        height=0.3\textwidth,
        xtick={0,0.1,...,0.7},
        xtick align=inside,
        ytick={0,200,...,1000}
        ]
    \addplot[fill=brown, draw=none] table [x=BinEdges, y=Frequency_DKT, col sep=comma] {background.csv};
    \addplot[fill=cyan, draw=none] table [x=BinEdges, y=Frequency_AKT, col sep=comma] {background.csv};
    \end{axis}
    \end{tikzpicture}
    \caption{The initial knowledge state distribution of students during the training process for ALPN-A2C and KT-KDM.}
    \label{figure:background}
\end{figure}

We conducted a statistical analysis on the initial knowledge levels of the $3000$ sampled students during the training process in ALPN-A2C and KT-KDM, and the results are depicted in Figure~\ref{figure:background}. We observed that, for the same group of students, the ALPN-A2C system using AKT could assess their knowledge background more comprehensively. On the other hand, the KT-KDM system using DKT provided similar evaluation results for the students. It means there is a significant difference in the standard deviations of the output distributions between the DKT and AKT. In other words, the complexity level varies when the two KT models are used as environments. Therefore, the KT-KDM system with DKT as the environment is able to converge faster during training, as its environment exhibits a lower complexity level. However, this also indicates that DKT struggles to estimate higher levels of knowledge, regardless of how well students perform in their learning activities.

Thus, our ALPN system incorporates AKT, allowing for a more nuanced assessment of student's knowledge levels. However, we must employ more exploratory DRL algorithms to tackle a more complex environment. In this regard, our proposed EPPO algorithm is the optimal choice.

\subsection{Learning Efficiency}

In addition to continuously providing effective learning paths for students with diverse knowledge backgrounds, students' learning time is also an important aspect to consider. Therefore, we examined whether the ALPN system could learn to recommend fewer learning materials during the training process while still helping students achieve the same learning goals.

\begin{figure}[h]
    \centering
    \begin{tikzpicture}
    \begin{axis}[
      xlabel={Episode},
      ylabel={Number of Attempts},
      xmin=0, xmax=3000,
      ymin=20, ymax=62,
      grid=major,
      legend entries={KT-KDM (DKT + A2C), ALPN-A2C (AKT + A2C), ALPN-PPO (AKT + PPO),\ ALPN-EPPO (AKT + EPPO)},
      legend style={at={(0.605,0.655)}, anchor=west, row sep=0.1cm},
      width=0.95\textwidth,
      height=0.4\textwidth,
      xtick={0,250,...,3000},
      ytick={20,25,...,60}
    ]
    \addplot[color=brown, mark=diamond*, mark repeat=125, line width=1.2pt] table [x expr=\coordindex, y=KT-KDM, col sep=comma] {effi.csv};
    \addplot[color=cyan, mark=triangle*, mark repeat=125, line width=1.2pt] table [x expr=\coordindex, y=ALPN-A2C, col sep=comma] {effi.csv};
    \addplot[color=blue, mark=*, mark repeat=125, line width=1.2pt] table [x expr=\coordindex, y=ALPN-PPO-noH, col sep=comma] {effi.csv};
    \addplot[color=magenta, mark=square*, mark repeat=125, line width=1.2pt] table [x expr=\coordindex, y=ALPN-PPO, col sep=comma] {effi.csv};
    \end{axis}
    \end{tikzpicture}
    \caption{The number of attempts students made to answer exercises in the learning path during the agent's training process.}
    \label{figure:effi}
\end{figure}
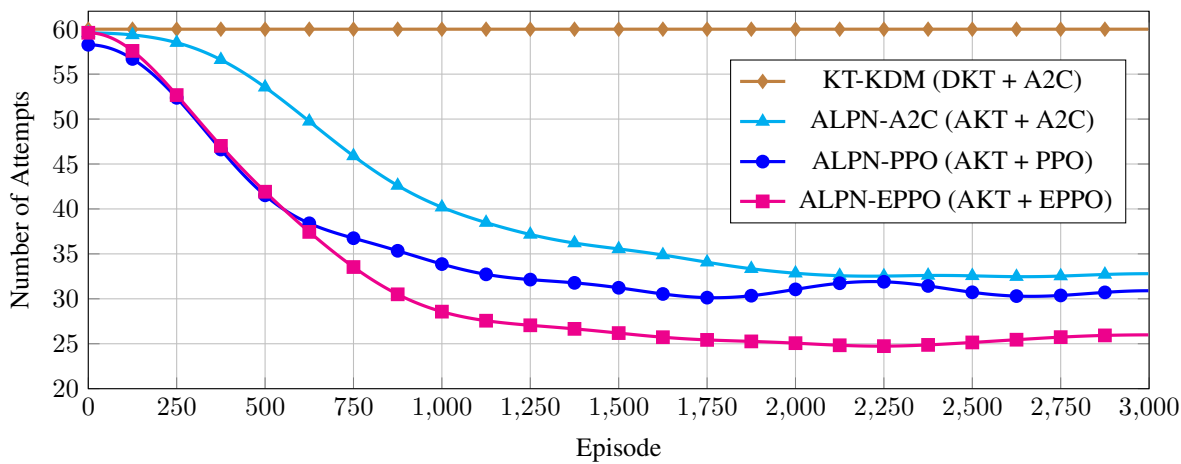

Figure~\ref{figure:effi} illustrates the variations in the length of learning paths generated by different systems as the number of training episodes progresses. It represents the changes in the number of learning activities students undertake to achieve their learning goals. From the figure, we observe that ALPN-EPPO enables students to achieve their learning goals with the fewest number of learning activities. It indicates that our proposed EPPO algorithm has a better impact on improving students' learning efficiency.

\subsection{Diversity}

In our ALPN system's reward function, we designed a penalty parameter $\lambda$ specifically to promote learning path diversity. This parameter assists the system in learning how to recommend appropriate learning materials for various knowledge states while maintaining diversity. A high level of diversity indicates a low repetition rate of learning paths. It indicates that the learning materials are recommended according to various knowledge states. Here, to evaluate the diversity in our ALPN system, we use the metric \textit{DIV} \cite{LDLP}:

\begin{equation}
\textit{DIV} = \dfrac{\Sigma_{i \neq j} \left(1-\dfrac{P_i \cap P_j}{l_i}\right)}{N - 1},
\end{equation}

\noindent where $P_i$ and $P_j$ represent the learning paths provided by the system to student $i$ and student $j$, respectively. $l_i$ denotes the length of learning path $P_i$, and $N$ represents the total number of learning paths. A higher \textit{DIV} value indicates a greater diversity in the learning paths provided by the system. We summarized the \textit{DIV} for each system in Table~\ref{div}, and observed that all ALPN systems exhibited higher diversity than KT-KDM, with ALPN-EPPO being the most prominent.

\begin{table}[!b]
\centering
\caption{Comparison among learning path planning methods.}
\label{div}
\setlength{\tabcolsep}{12pt}
\begin{tabular}{cccccc}
    \toprule
    Methods & KT-KDM & ALPN-A2C & ALPN-PPO & \textbf{ALPN-EPPO} \\
    \midrule
    Diversity ($\textit{DIV}$) & 0.881 & 0.952 & 0.965 & \textbf{0.968} \\
    \bottomrule
\end{tabular}
\end{table}



\begin{figure}[!b]
\centering
\begin{tikzpicture}
\node[draw=none] (n) { };
\node[circle, fill=area_0, text=white, minimum size=10pt, align=center, inner sep=2pt, draw=gray, line width=1.5pt, right=0.3cm of n] (n0) { };
\node[circle, fill=area_1, text=white, minimum size=10pt, align=center, inner sep=2pt, draw=gray, line width=1.5pt, right=1.4cm of n0] (n1) { };
\node[circle, fill=area_2, text=white, minimum size=10pt, align=center, inner sep=2pt, draw=gray, line width=1.5pt, right=1.75cm of n1] (n2) { };
\node[circle, fill=area_3, text=white, minimum size=10pt, align=center, inner sep=2pt, draw=gray, line width=1.5pt, right=1.7cm of n2] (n3) { };
\node[circle, fill=area_4, text=white, minimum size=10pt, align=center, inner sep=2pt, draw=gray, line width=1.5pt, right=2.8cm of n3] (n4) { };
\node[circle, fill=area_5, text=white, minimum size=10pt, align=center, inner sep=2pt, draw=gray, line width=1.5pt, right=3.0cm of n4] (n5) { };
\node[star, fill=brickred, text=white, minimum size=10pt, align=center, inner sep=2pt, right=1.5cm of n5] (end) { };

\node[anchor=west, font=\footnotesize] at (n0.east) (n00) {algebra \qquad \;\ arithmetic \qquad \;\ geometry \qquad \,\, analytic-geometry \qquad \; probability-statistics \qquad\, calculus \qquad\ end};
\end{tikzpicture}

\vspace{5pt}

\begin{tikzpicture}
    \begin{axis}[
        xlabel={Timestep ($t$)},
        ylabel={Knowledge level ($\textit{APR}_t$)},
        xmin=-1, xmax=26,
        ymin=0.1, ymax=1,
        grid=both,
        grid style={line width=.1pt, draw=gray!10},
        major grid style={line width=.2pt, draw=gray!50},
        width=1.0\textwidth,
        height=0.4\textwidth,
        legend entries={\ Student $\textit{A}$, \ Student $\textit{B}$},
        legend style={at={(0.807, 0.28)}, anchor=west, row sep=0.1cm},
        xtick={0, 1, ..., 26},
        ytick={0.2, 0.3, ..., 1},
    ]

    \addplot+[color=gray, mark=none, dotted, line width=1.5pt] coordinates {(0, 0.303) (1, 0.300) (2, 0.349) (3, 0.557) (4, 0.531) (5, 0.632) (6, 0.590) (7, 0.711) (8, 0.766) (9, 0.684) (10, 0.748) (11, 0.700) (12, 0.785) (13, 0.837) (14, 0.828) (15, 0.853) (16, 0.867) (17, 0.867) (18, 0.870) (19, 0.896) (20, 0.849) (21, 0.899) (22, 0.901)};
    \node[circle, fill=area_3, text=white, minimum size=10pt, align=center, inner sep=2pt, font=\scriptsize, draw=gray, line width=1.5pt] at (axis cs:0, 0.303) { };
    \node[circle, fill=area_2, text=white, minimum size=10pt, align=center, inner sep=2pt, font=\scriptsize, draw=gray, line width=1.5pt] at (axis cs:1, 0.300) { };
    \node[circle, fill=area_2, text=white, minimum size=10pt, align=center, inner sep=2pt, font=\scriptsize, draw=gray, line width=1.5pt] at (axis cs:2, 0.349) { };
    \node[circle, fill=area_3, text=white, minimum size=10pt, align=center, inner sep=2pt, font=\scriptsize, draw=gray, line width=1.5pt] at (axis cs:3, 0.557) { };
    \node[circle, fill=area_4, text=white, minimum size=10pt, align=center, inner sep=2pt, font=\scriptsize, draw=gray, line width=1.5pt] at (axis cs:4, 0.531) { };
    \node[circle, fill=area_2, text=white, minimum size=10pt, align=center, inner sep=2pt, font=\scriptsize, draw=gray, line width=1.5pt] at (axis cs:5, 0.632) { };
    \node[circle, fill=area_1, text=white, minimum size=10pt, align=center, inner sep=2pt, font=\scriptsize, draw=gray, line width=1.5pt] at (axis cs:6, 0.590) { };
    \node[circle, fill=area_0, text=white, minimum size=10pt, align=center, inner sep=2pt, font=\scriptsize, draw=gray, line width=1.5pt] at (axis cs:7, 0.711) { };
    \node[circle, fill=area_2, text=white, minimum size=10pt, align=center, inner sep=2pt, font=\scriptsize, draw=gray, line width=1.5pt] at (axis cs:8, 0.766) { };
    \node[circle, fill=area_0, text=white, minimum size=10pt, align=center, inner sep=2pt, font=\scriptsize, draw=gray, line width=1.5pt] at (axis cs:9, 0.684) { };
    \node[circle, fill=area_0, text=white, minimum size=10pt, align=center, inner sep=2pt, font=\scriptsize, draw=gray, line width=1.5pt] at (axis cs:10, 0.748) { };
    \node[circle, fill=area_1, text=white, minimum size=10pt, align=center, inner sep=2pt, font=\scriptsize, draw=gray, line width=1.5pt] at (axis cs:11, 0.700) { };
    \node[circle, fill=area_0, text=white, minimum size=10pt, align=center, inner sep=2pt, font=\scriptsize, draw=gray, line width=1.5pt] at (axis cs:12, 0.785) { };
    \node[circle, fill=area_0, text=white, minimum size=10pt, align=center, inner sep=2pt, font=\scriptsize, draw=gray, line width=1.5pt] at (axis cs:13, 0.837) { };
    \node[circle, fill=area_4, text=white, minimum size=10pt, align=center, inner sep=2pt, font=\scriptsize, draw=gray, line width=1.5pt] at (axis cs:14, 0.828) { };
    \node[circle, fill=area_5, text=white, minimum size=10pt, align=center, inner sep=2pt, font=\scriptsize, draw=gray, line width=1.5pt] at (axis cs:15, 0.853) { };
    \node[circle, fill=area_1, text=white, minimum size=10pt, align=center, inner sep=2pt, font=\scriptsize, draw=gray, line width=1.5pt] at (axis cs:16, 0.867) { };
    \node[circle, fill=area_1, text=white, minimum size=10pt, align=center, inner sep=2pt, font=\scriptsize, draw=gray, line width=1.5pt] at (axis cs:17, 0.867) { };
    \node[circle, fill=area_1, text=white, minimum size=10pt, align=center, inner sep=2pt, font=\scriptsize, draw=gray, line width=1.5pt] at (axis cs:18, 0.870) { };
    \node[circle, fill=area_0, text=white, minimum size=10pt, align=center, inner sep=2pt, font=\scriptsize, draw=gray, line width=1.5pt] at (axis cs:19, 0.896) { };
    \node[circle, fill=area_3, text=white, minimum size=10pt, align=center, inner sep=2pt, font=\scriptsize, draw=gray, line width=1.5pt] at (axis cs:20, 0.879) { };
    \node[circle, fill=area_0, text=white, minimum size=10pt, align=center, inner sep=2pt, font=\scriptsize, draw=gray, line width=1.5pt] at (axis cs:21, 0.899) { };
    \node[star, fill=brickred, text=white, minimum size=10pt, align=center, inner sep=2pt, font=\scriptsize] at (axis cs:22, 0.901) { };


    \addplot+[color=gray, mark=none, line width=1.5pt] coordinates {(0, 0.291) (1, 0.223) (2, 0.379) (3, 0.388) (4, 0.274) (5, 0.304) (6, 0.309) (7, 0.346) (8, 0.463) (9, 0.515) (10, 0.688) (11, 0.726) (12, 0.726) (13, 0.777) (14, 0.801) (15, 0.808) (16, 0.820) (17, 0.815) (18, 0.821) (19, 0.843) (20, 0.853) (21, 0.831) (22, 0.818) (23, 0.839) (24, 0.855) (25, 0.901)};
    \node[circle, fill=area_0, text=white, minimum size=10pt, align=center, inner sep=2pt, font=\scriptsize, draw=gray, line width=1.5pt] at (axis cs:0, 0.281) { };
    \node[circle, fill=area_0, text=white, minimum size=10pt, align=center, inner sep=2pt, font=\scriptsize, draw=gray, line width=1.5pt] at (axis cs:1, 0.223) { };
    \node[circle, fill=area_4, text=white, minimum size=10pt, align=center, inner sep=2pt, font=\scriptsize, draw=gray, line width=1.5pt] at (axis cs:2, 0.379) { };
    \node[circle, fill=area_1, text=white, minimum size=10pt, align=center, inner sep=2pt, font=\scriptsize, draw=gray, line width=1.5pt] at (axis cs:3, 0.388) { };
    \node[circle, fill=area_0, text=white, minimum size=10pt, align=center, inner sep=2pt, font=\scriptsize, draw=gray, line width=1.5pt] at (axis cs:4, 0.274) { };
    \node[circle, fill=area_1, text=white, minimum size=10pt, align=center, inner sep=2pt, font=\scriptsize, draw=gray, line width=1.5pt] at (axis cs:5, 0.304) { };
    \node[circle, fill=area_4, text=white, minimum size=10pt, align=center, inner sep=2pt, font=\scriptsize, draw=gray, line width=1.5pt] at (axis cs:6, 0.309) { };
    \node[circle, fill=area_1, text=white, minimum size=10pt, align=center, inner sep=2pt, font=\scriptsize, draw=gray, line width=1.5pt] at (axis cs:7, 0.346) { };
    \node[circle, fill=area_0, text=white, minimum size=10pt, align=center, inner sep=2pt, font=\scriptsize, draw=gray, line width=1.5pt] at (axis cs:8, 0.463) { };
    \node[circle, fill=area_2, text=white, minimum size=10pt, align=center, inner sep=2pt, font=\scriptsize, draw=gray, line width=1.5pt] at (axis cs:9, 0.515) { };
    \node[circle, fill=area_0, text=white, minimum size=10pt, align=center, inner sep=2pt, font=\scriptsize, draw=gray, line width=1.5pt] at (axis cs:10, 0.688) { };
    \node[circle, fill=area_4, text=white, minimum size=10pt, align=center, inner sep=2pt, font=\scriptsize, draw=gray, line width=1.5pt] at (axis cs:11, 0.726) { };
    \node[circle, fill=area_0, text=white, minimum size=10pt, align=center, inner sep=2pt, font=\scriptsize, draw=gray, line width=1.5pt] at (axis cs:12, 0.726) { };
    \node[circle, fill=area_1, text=white, minimum size=10pt, align=center, inner sep=2pt, font=\scriptsize, draw=gray, line width=1.5pt] at (axis cs:13, 0.777) { };
    \node[circle, fill=area_1, text=white, minimum size=10pt, align=center, inner sep=2pt, font=\scriptsize, draw=gray, line width=1.5pt] at (axis cs:14, 0.801) { };
    \node[circle, fill=area_0, text=white, minimum size=10pt, align=center, inner sep=2pt, font=\scriptsize, draw=gray, line width=1.5pt] at (axis cs:15, 0.808) { };
    \node[circle, fill=area_4, text=white, minimum size=10pt, align=center, inner sep=2pt, font=\scriptsize, draw=gray, line width=1.5pt] at (axis cs:16, 0.820) { };
    \node[circle, fill=area_1, text=white, minimum size=10pt, align=center, inner sep=2pt, font=\scriptsize, draw=gray, line width=1.5pt] at (axis cs:17, 0.815) { };
    \node[circle, fill=area_3, text=white, minimum size=10pt, align=center, inner sep=2pt, font=\scriptsize, draw=gray, line width=1.5pt] at (axis cs:18, 0.821) { };
    \node[circle, fill=area_2, text=white, minimum size=10pt, align=center, inner sep=2pt, font=\scriptsize, draw=gray, line width=1.5pt] at (axis cs:19, 0.843) { };   
    \node[circle, fill=area_0, text=white, minimum size=10pt, align=center, inner sep=2pt, font=\scriptsize, draw=gray, line width=1.5pt] at (axis cs:20, 0.853) { };
    \node[circle, fill=area_0, text=white, minimum size=10pt, align=center, inner sep=2pt, font=\scriptsize, draw=gray, line width=1.5pt] at (axis cs:21, 0.831) { };
    \node[circle, fill=area_1, text=white, minimum size=10pt, align=center, inner sep=2pt, font=\scriptsize, draw=gray, line width=1.5pt] at (axis cs:22, 0.818) { };
    \node[circle, fill=area_1, text=white, minimum size=10pt, align=center, inner sep=2pt, font=\scriptsize, draw=gray, line width=1.5pt] at (axis cs:23, 0.839) { };
    \node[circle, fill=area_4, text=white, minimum size=10pt, align=center, inner sep=2pt, font=\scriptsize, draw=gray, line width=1.5pt] at (axis cs:24, 0.855) { };
    \node[star, fill=brickred, text=white, minimum size=10pt, align=center, inner sep=2pt, font=\scriptsize] at (axis cs:25, 0.901) { };
    \end{axis}
\end{tikzpicture}

\vspace{-5pt}

\caption{The students' learning paths with similar initial knowledge levels.}
\label{figure:paths}
\end{figure}
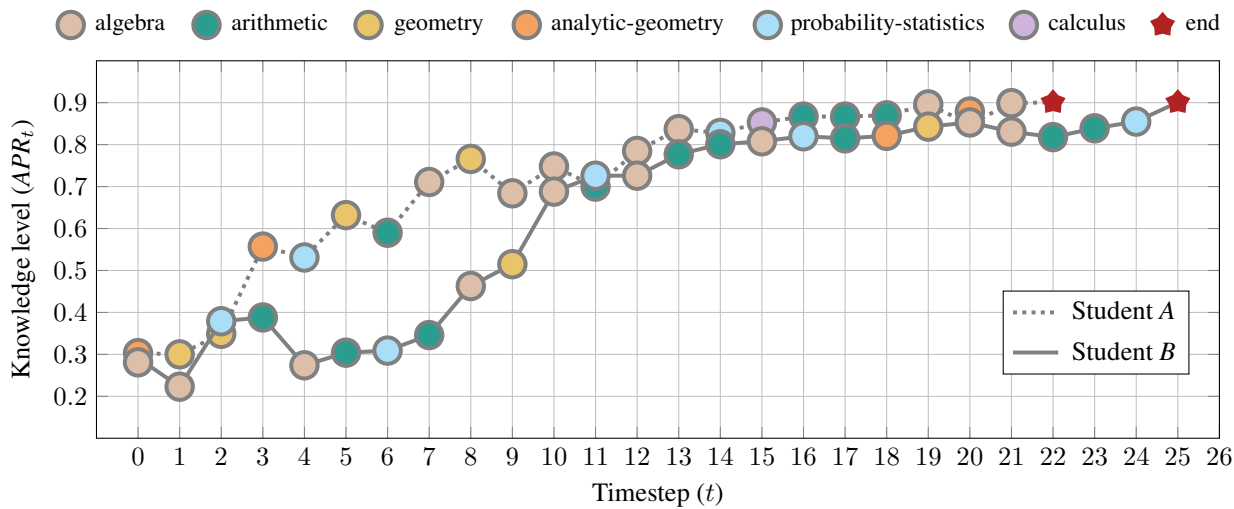

We also examined whether the ALPN system would provide different learning paths for students with similar knowledge levels. It also reflects that the system can generate learning paths with high diversity. We use the trained ALPN-EPPO to sample two students with very close initial knowledge levels and offer them individual learning paths. Figure~\ref{figure:paths} display the result of the comparison between the learning paths of the two students. The figure illustrates the changes in knowledge levels for student $\textit{A}$ and student $\textit{B}$ along their respective learning paths. The nodes on the graph represent the learning materials, i.e., the exercises, that students received at that timestep. We use pre-defined categories from the Junyi dataset to annotate each exercise with a color corresponding to its area. Red asterisks indicate the timesteps at which students achieve learning goals. From this figure, even when these two students have a very similar initial knowledge level, there are still specific differences in the learning paths they receive. Moreover, both students can successfully achieve their learning goals. It indicates that our ALPN-EPPO system helps students achieve their goals and exhibits diversity in generating learning paths.

\subsection{Cumulative Reward}

\begin{figure}[!b]
\centering
\begin{tikzpicture}
\begin{axis}[
  xlabel={Episode},
  ylabel={Cumulative reward},
  xmin=0, xmax=3000,
  ymin=-500, ymax=9500,
  grid=major,
  legend entries={\ ALPN-EPPO, ALPN-PPO, ALPN-A2C},
  legend style={at={(0.77,0.25)}, anchor=west, row sep=0.1cm},
  width=0.95\textwidth,
  height=0.4\textwidth,
  xtick={0,250,...,3000},
  ytick={0,1500,...,9000},
  yticklabel style={/pgf/number format/fixed, /pgf/number format/precision=0},
  scaled y ticks=false,
]
\addplot[color=magenta, mark=square*, mark repeat=125, line width=1.2pt] table [x expr=\coordindex, y=ALPN-EPPO, col sep=comma] {reward.csv};
\addplot[color=blue, mark=*, mark repeat=125, line width=1.2pt] table [x expr=\coordindex, y=ALPN-PPO, col sep=comma] {reward.csv};
\addplot[color=cyan, mark=triangle*, mark repeat=125, line width=1.2pt] table [x expr=\coordindex, y=ALPN-A2C, col sep=comma] {reward.csv};
\end{axis}
\end{tikzpicture}
\caption{The cumulative reward during the agents' training process.}
\label{figure:reward}
\end{figure}
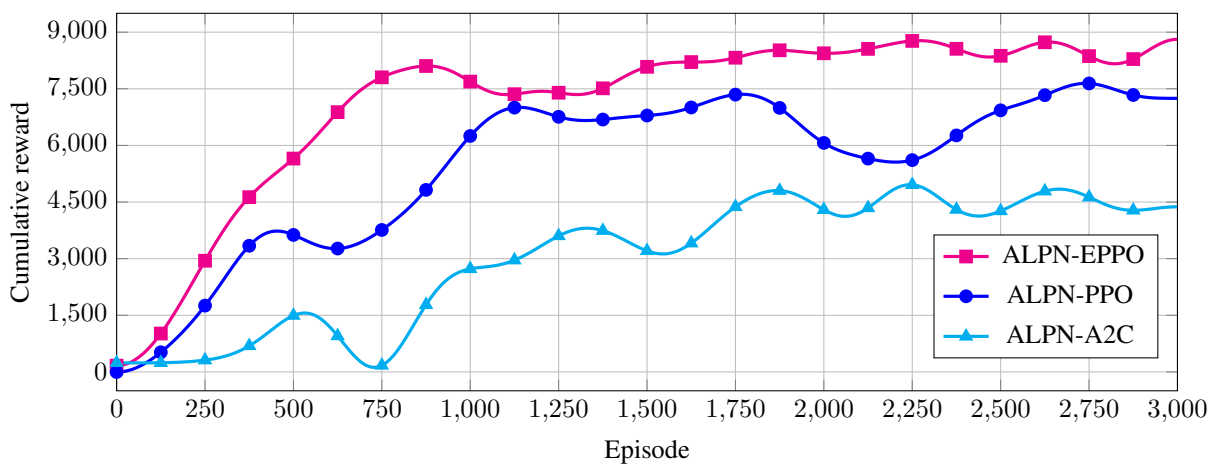

We can utilize the cumulative reward obtained during training by the system as another important evaluation metric because it represents the system's overall performance in an environment. Figure~\ref{figure:reward} illustrates the changes in cumulative reward during the training process for our ALPN system using EPPO, PPO, and A2C, respectively. We can observe that ALPN-EPPO achieves the highest cumulative reward, indicating the outstanding performance of EPPO in our learning path recommendation task. Furthermore, since our reward function differs from the one used in the KT-KDM system, we do not compare it.

\subsection{Learning Process}

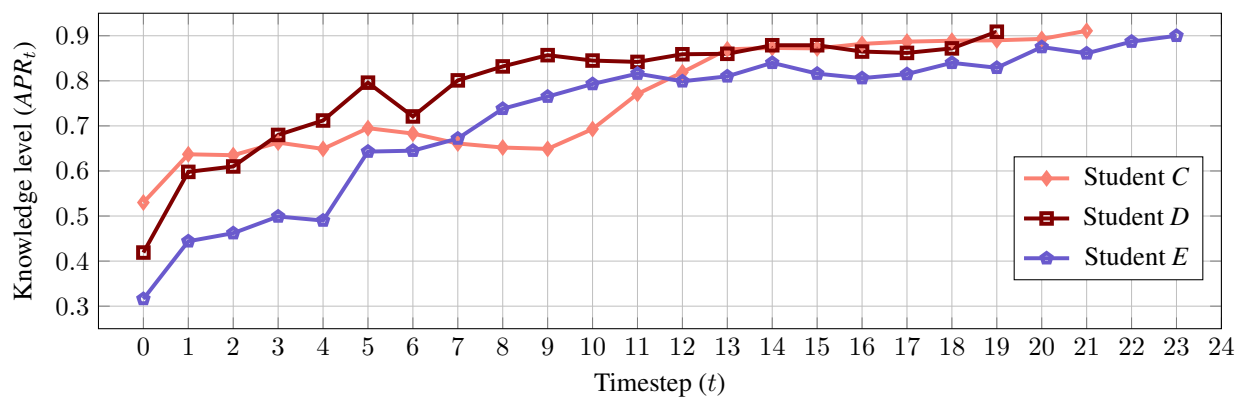
\begin{figure}[!b]
\centering
\begin{tikzpicture}
    \begin{axis}[
        xlabel={Timestep ($t$)},
        ylabel={Knowledge level ($\textit{APR}_t$)},
        xmin=-1, xmax=24,
        ymin=0.25, ymax=0.95,
        grid=both,
        grid style={line width=.1pt, draw=gray!10},
        major grid style={line width=.2pt, draw=gray!50},
        width=1.0\textwidth,
        height=0.35\textwidth,
        legend entries={\ Student $\textit{C}$,\ Student $\textit{D}$,\ Student $\textit{E}$},
        legend style={at={(0.815, 0.35)}, anchor=west, row sep=0.1cm},
        xtick={0, 1, ..., 24},
        ytick={0.2, 0.3, ..., 0.95},
    ]

    \addplot+[color=salmon, mark=diamond, line width=1.5pt] coordinates {(0, 0.530) (1, 0.637) (2, 0.635) (3, 0.663) (4, 0.649) (5, 0.695) (6, 0.683) (7, 0.661) (8, 0.652) (9, 0.649) (10, 0.693) (11, 0.771) (12, 0.819) (13, 0.870) (14, 0.873) (15, 0.872) (16, 0.882) (17, 0.887) (18, 0.889) (19, 0.890) (20, 0.893) (21, 0.911)};

    \addplot+[color=maroon, mark=square, line width=1.5pt] coordinates {(0, 0.419) (1, 0.598) (2, 0.610) (3, 0.680) (4, 0.712) (5, 0.796) (6, 0.721) (7, 0.801) (8, 0.832) (9, 0.857) (10, 0.845) (11, 0.842) (12, 0.859) (13, 0.860) (14, 0.879) (15, 0.879) (16, 0.865) (17, 0.862) (18, 0.872) (19, 0.909)};

    \addplot+[color=slateblue, mark=pentagon, line width=1.5pt] coordinates {(0, 0.316) (1, 0.444) (2, 0.462) (3, 0.499) (4, 0.490) (5, 0.643) (6, 0.645) (7, 0.672) (8, 0.738) (9, 0.765) (10, 0.793) (11, 0.816) (12, 0.799) (13, 0.810) (14, 0.840) (15, 0.816) (16, 0.806) (17, 0.815) (18, 0.840) (19, 0.829) (20, 0.875) (21, 0.861) (22, 0.887) (23, 0.900)};



\end{axis}
\end{tikzpicture}
\vspace{-15pt}
\caption{The evolution of students' knowledge level over time.}
\label{figure:levels}
\end{figure}

Next, we sampled three additional students and used the ALPN-EPPO system to provide them with learning paths. The changes in their knowledge levels during the learning process are depicted in Figure~\ref{figure:levels}. The graph shows that not every student's knowledge level consistently increases. For instance, student $\textit{C}$'s knowledge level fluctuates in the first half of the learning process. It occurs because our system considers correct and incorrect responses during exercises. When students frequently answer incorrectly, their knowledge state naturally does not improve. However, our EPPO adjusts after a student answers multiple exercises incorrectly, recommending exercises that better match the student's current level. This adaptability to the student's state allows our ALPN-EPPO system to assist each student in achieving their learning goals.

\begin{figure}[t]
  \begin{center}
    \begin{tikzpicture}
    \end{tikzpicture}

    \begin{minipage}{0.05\textwidth}
      \begin{tikzpicture}[every node/.style={font=\footnotesize, text width=3cm, align=right}]
          \node(nodeGap0) {};
          \node[below of=nodeGap0, node distance=0.25cm] (node1) {algebra};
          \node[below of=node1, node distance=0.5cm] (node2) {analytic-geometry};
          \node[below of=node2, node distance=0.05cm] (nodeGap00) {};
          \node[below of=nodeGap00, node distance=0.5cm] (node3) {arithmetic};
          \node[below of=node3, node distance=0.03cm] (nodeGap1) {};
          \node[below of=nodeGap1, node distance=0.5cm] (node4) {calculus};
          \node[below of=node4, node distance=0.07cm] (nodeGap2) {};
          \node[below of=nodeGap2, node distance=0.5cm] (node5) {geometry};
          \node[below of=node5, node distance=0.03cm] (nodeGap3) {};
          \node[below of=nodeGap3, node distance=0.5cm] (node6) {logics};
          \node[below of=node6, node distance=0.5cm] (node7) {probability-statistics};
      \end{tikzpicture}
    \end{minipage}
    \hfill
    \begin{minipage}{0.84\textwidth}
      \centering
      \begin{tabular}{*{33}{R}}
        0.35&0.39&0.3&0.26&0.35&0.49&0.43&0.42&0.47&0.42&0.5&0.6&0.52&0.5&0.66&0.61&0.56&0.61&0.66&0.63&0.52&0.65&0.65&0.56&0.54&0.61&0.66&0.7&0.7&0.73&0.73&0.76&0.78 \\
        0.39&0.32&0.3&0.25&0.32&0.45&0.33&0.43&0.46&0.41&0.58&0.71&0.65&0.65&0.74&0.7&0.69&0.73&0.74&0.65&0.55&0.6&0.6&0.52&0.59&0.7&0.75&0.75&0.75&0.8&0.76&0.79&0.82 \\
        0.4&0.4&0.34&0.32&0.38&0.47&0.38&0.39&0.42&0.39&0.5&0.65&0.58&0.61&0.74&0.67&0.65&0.71&0.76&0.69&0.6&0.73&0.74&0.69&0.69&0.72&0.76&0.8&0.8&0.84&0.84&0.85&0.86 \\
        0.24&0.34&0.2&0.21&0.23&0.25&0.3&0.29&0.37&0.39&0.42&0.58&0.44&0.5&0.44&0.34&0.25&0.31&0.38&0.49&0.38&0.36&0.4&0.49&0.58&0.59&0.58&0.58&0.6&0.63&0.63&0.62&0.63 \\
        0.42&0.38&0.31&0.29&0.37&0.42&0.36&0.38&0.45&0.42&0.51&0.6&0.49&0.53&0.67&0.61&0.59&0.63&0.65&0.59&0.52&0.63&0.61&0.57&0.6&0.66&0.68&0.71&0.71&0.75&0.73&0.75&0.77 \\
        0.53&0.56&0.4&0.46&0.42&0.54&0.3&0.39&0.42&0.43&0.39&0.68&0.59&0.59&0.71&0.65&0.51&0.64&0.59&0.54&0.53&0.52&0.47&0.43&0.71&0.74&0.8&0.83&0.8&0.79&0.78&0.78&0.79 \\
        0.31&0.33&0.26&0.25&0.31&0.46&0.36&0.31&0.36&0.32&0.4&0.55&0.51&0.51&0.7&0.63&0.6&0.66&0.61&0.6&0.53&0.58&0.6&0.52&0.57&0.64&0.67&0.7&0.68&0.66&0.68&0.65&0.67 \\
      \end{tabular}
    \end{minipage}
    \vspace{5pt}
    \caption{The evolution of a student's mastery level of each area during the learning process.}
    \label{figure:heat}
  \end{center}
\end{figure}
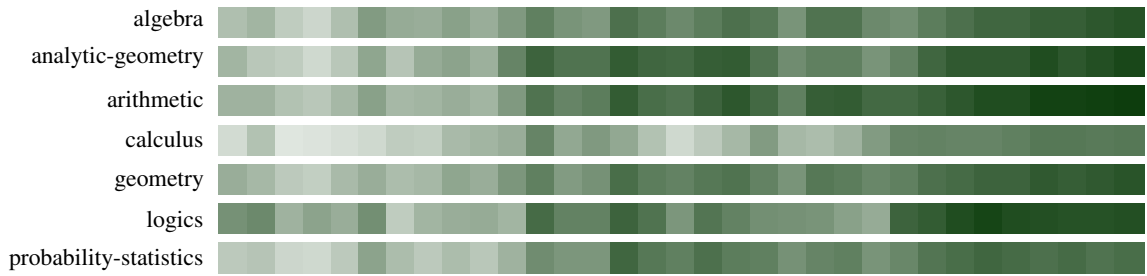

We also examined the changes in the knowledge state of a single students=, i.e., the mastery level of various knowledge components. When students aim to maximize their learning outcome towards $1$, they must maximize their mastery level of each knowledge component. On the other hand, when a student's learning goals are more modest (e.g., $\beta = 0.8$), they can achieve their goals by improving the mastery of only a subset of concepts. However, we expect that after students have completed the learning path provided by the system, they will become more proficient in each knowledge. To verify this, we present in Figure~\ref{figure:heat} the proficiency changes of a student for each area throughout the learning process. The figure, from left to right, represents the student's learning process. Each vertical column represents the student's proficiency level for each area at a specific moment, with darker colors indicating higher proficiency. The figure shows that after completing this learning process, the student has significantly improved in every area rather than just a few.

\section{Conclusion}

This paper presented an innovative approach to enhance E-learning systems through the Adaptive Learning Path Navigation (ALPN) system. By integrating the Attentive Knowledge Tracing (AKT) model with the newly proposed Entropy-enhanced version of Proximal Policy Optimization (EPPO), the ALPN system generates adaptive learning paths tailored to students' knowledge states, thereby increasing the effectiveness of online learning.

The application of AKT, which employs an attention mechanism, accurately estimates a student's knowledge state. The EPPO model, on the other hand, optimizes the recommendation of learning materials, successfully outperforming the traditional PPO model in our task by enhancing exploratory capabilities. The ALPN system harmonizes these two models to facilitate highly adaptive and effective learning paths, offering the promising potential for adaptive E-learning systems.

The performance of the ALPN system was tested against the existing Knowledge Tracing based Knowledge Demand Model (KT-KDM). Our system demonstrated significant improvements, outperforming the KT-KDM method by 8.2\% on average to maximize students' learning outcomes. Additionally, the ALPN system displayed superior diversity in generating learning paths, with a 10.5\% higher rate than the KT-KDM.

Even with the significant achievements demonstrated by the ALPN system, future work should further refine the proposed approach. These improvements include tailoring the system to cater to a broader range of learning styles, optimizing the model to increase further learning outcome improvements, and enhancing the model's scalability to accommodate even more significant numbers of students and diverse learning materials. Furthermore, exploring additional personalization parameters, such as learning style preferences or motivational factors, may prove beneficial in creating even more tailored learning paths.

In conclusion, integrating AKT and EPPO in the proposed ALPN system marked a significant advancement in adaptive E-learning. The system's ability to generate reliable and adaptive learning paths enhances learning efficiency, making E-learning an even more effective educational tool in the digital era. By continually seeking to refine and improve such models, the education sector can stay at the forefront of technological advancements, ensuring quality learning experiences for all students.

\bibliographystyle{unsrt}  
\bibliography{references}  

\end{document}